# A multi-temporal dataset for mapping burned areas in the Brazilian Cerrado using time series of remote sensing imagery

Alisson Cleiton de Oliveira [a], Thales Sehn Körting [b]

[a b] Earth Observation and Geoinformatics Division (DIOTG), National Institute for Space Research (INPE), São José dos Campos, SP, Brazil
[a] alisson.oliveira@inpe.br (ORCID ID: 0000-0001-5177-3730), [b] thales.korting@inpe.br (ORCID ID: 0000-0002-0876-0501)

## ABSTRACT

This paper introduces a multi-temporal tabular dataset derived from satellite images to map burned areas in the Chapada dos Veadeiros National Park, Goiás, Brazil, for 2020–2022. The dataset contains blue, green, red, and near-infrared bands, as well as the BAI, EVI, GEMI, NDVI, and NDWI spectral indices from the WFI sensor on the CBERS-4A, CBERS-4, and AMAZONIA-1 satellites, organized into a regular grid. We applied the Random Forest classifier to develop and validate models based on samples labeled as totally burned, partially burned, and non-burned. Two classification approaches were tested: one combining burned and non-burned areas into binary classes, and another distinguishing between totally burned (TB), partially burned (PB), and non-burned (NB) classes. Seven validation approaches assessed different post-classification combinations, focusing on accuracy, precision, recall, and intersection over union (IoU) metrics. Results showed higher IoU when TB, PB, and NB were used as individual classes and TB was reclassified as burned area (BA) while PB and NB were grouped as non-burned. Comparing the annual results of this approach to the MCD64A1 product, the errors of omission for the BA class were 22% in 2020, 28% in 2021 and 59% in 2022, while the errors of commission were 46%, 43% and 46% respectively. The study highlights the utility of the WFI sensor for burned area mapping without inter-satellite spectral calibration and suggests further exploration with other machine learning algorithms to evaluate the dataset potential and limitations.

**KEYWORDS**: Satellite images. WFI sensor. CBERS-4. CBERS-4. AMAZONIA-1. Random Forest.

CONTACT Alisson Cleiton de Oliveira, alisson.oliveira@inpe.br, Earth Observation and Geoinformatics Division (DIOTG), National Institute for Space Research (INPE), São José dos Campos, SP, Brazil

## 1. Introduction

From a Remote Sensing (RS) perspective, modifications caused by forest fires can be understood as abrupt and non-permanent changes in spectral response from vegetation cover (Milne, 1988, Coppin et al, 2004; Rossi; Santos, 2020; Shimabukuro et al., 2020). RS enables the decoding of electromagnetic radiation and the conversion of radiation emitted or reflected from targets on the Earth's surface into images based on the spectral behavior and characteristics of the targets (Jensen, 2007). RS also enables several analyzes related to the Earth's resources and climate (Szpakowski & Jensen, 2019). As land use practices related to fire become more intense, and the incidence of human-caused fires increases (Pivello et al., 2021), RS applications play an important role in understanding the historical and current dynamics of these events and in assessing the impacts of fires at different scales of analysis, as well as for near-real time monitoring (Chuvieco et al., 2019).

As explained by Archibald and Roy (2009), there are two main types of fire-related products processed with RS-based data: heat-related products (active fire or fire-foci) and products that focus on the vegetation biophysical changes observed after fires (burned areas). It is important to emphasize the spectral distinction between active fires, which is the direct product of combustion (Butler et al., 2004), and burned areas, which is, according to Lentile et al. (2006), the simplest remote measure of post-fire effects. Along the electromagnetic spectrum, specific regions contain information about the spectral behavior of these different targets, and while the energy released by fires can be detected in the infrared range and indicates active fires, the electromagnetic energy from burned surfaces can also be provided by the solar energy reflected by these affected areas and can be detected over a broader spectral range (Chuvieco et al., 2019), including the visible region, despite the possibility of using pre and post-fire images or more dense time series to detect the location and extent of this target (Liu et al., 2023).

The Brazilian Cerrado is one of the most species-rich savannas in the world and a fire-dependent biome. Natural fires have been present for at least 4 million years, and fire has been a natural selection factor that has led to the flora of this biome being characterized today as fire-resistant due to morphological and physiological adaptations (Nascimento, 2001). However, it is important to highlight that natural fires occur between 3 and 6 years according to specific fire regimes (Ramos-Neto & Pivello, 2000; Júnior et al., 2014). Despite this evolutionary relationship, the current occurrence of human-caused fires has a negative impact on the ecosystems of the Cerrado (Durigan et al., 2020; Fidelis et al., 2018), as these fires occur at unfavorable seasons of the year, last longer, are more intense and threaten the biodiversity of the Cerrado, which cannot withstand the high frequency of these events and the high temperatures that the flames can reach.

The analysis of spectral characteristics of burned areas is limited when a sensor, such as the WFI (Wide Field Imager) on board CBERS-4 (China-Brazil Earth Resources Satellite), CBERS-4A and AMAZONIA-1 satellites, operates in the BGR NIR (Blue, Green, Red and Near-Infrared) bands. However, the integration of data from these three orbital platforms enables Brazil to have a constellation of satellites equipped with WFI sensors, capable of providing images with a spatial resolution of 55 m (CBERS-4A) and 64 m (CBERS-4 and AMAZONIA-1) at intervals from 1 to 3 days (Oldoni, 2022). As of the present moment, there has been no systematic mapping of burned areas using these satellites.

Such kind of data is promising for artificial intelligence (AI) algorithms that deal with pattern recognition in time series. The ability to work with large volumes of data and recognize patterns are key advantages of using AI, and this is important in RS as orbital platforms are constantly generating new data for Earth observation. In recent years, machine learning (ML) techniques have been widely used in environmental sciences (Stroppiana et al., 2021), with applications in mapping burned areas (Wood, 2021). According to Tiwari et al. (2018), ML enables computer systems to learn from examples in the form of data. There are two main categories of ML approaches, corresponding to “unsupervised learning” and “supervised learning”. The first approach is more unpredictable, while supervised learning allows for better expectations of results when a model has been trained based on human experience (Liu et al., 2018). It is also possible to use hybrid approaches based on the previous categories (Mehmood et al., 2022).

Since the 1980s, RS has been increasingly used to map burned areas, with advancements in satellite-based products enabling global monitoring of fire-affected regions (Mouillot et al., 2014). Global products like the MODIS Direct Broadcast Monthly Burned Area Product Collection 6 (MCD64A1) derived from the Moderate Resolution Imaging Spectrometer (MODIS) sensor provide monthly burned area coverage, while regional products such as AQM1km focus on Brazil. These products rely on hybrid algorithms that combine time series of surface reflectance data and active fire data to improve detection across different vegetation types and regions (Giglio et al., 2018). However, challenges remain in complex vegetation regions like the Brazilian Cerrado, where local vegetation characteristics can introduce uncertainties (Rodrigues et al., 2019).

In Brazil, initiatives such as the MapBiomas project have mapped burned areas from 1985 to 2022 using Landsat imagery and deep learning (DL) (Alencar et al., 2022). The project regularly updates its methodology to improve accuracy, with variations in burned area maps resulting from different algorithms and analysis scales (Pessôa et al., 2020; Shimabukuro et al., 2020). To ensure the reliability of these products, validation with independent reference data, often involving human image interpretation, helps reduce errors and refine assessment quality (Humber et al., 2019). Other approaches outside Brazil employ different DL algorithms, but Random Forest (RF) is still used due to its reproducibility and robustness, as recently exemplified by Bastarrika et al. (2024), who developed an automatic method for mapping burned areas using Sentinel-2 imagery, and Visible Infrared Imaging Radiometer Suite (VIIRS) and MODIS fire foci.

RF is a non-parametric supervised algorithm based on decision trees for regression and classification tasks. In this classifier, two main parameters are fixed a priori: the number of trees and the percentage of classification and validation samples. The selection of these samples from the dataset specified by the user is random and therefore the algorithm may use different subtraining sets at each round (Breiman, 2001; Kulkarni & Lowe, 2016; Belgiu & Drăguţ, 2016), i.e. for the construction of each tree. RS-based data rarely satisfy the assumption of a normal spectral distribution (Lu & Weng, 2007). In such cases, non-parametric classifiers are often used (Belgiu & Drăguţ, 2016). To generate a RS image classification applying random forests, the first node of a decision tree starts with the entire set of samples, which is then divided into training and validation subsets. For time series classification of geographical data, a dataset can be understood as a data cube, where each attribute can be interpreted as an observed day or a temporal metric extracted from multiple days, depending on the method used to generate the data cube (Körting et al., 2013; Vieira et al., 2022).

A RS-based method for mapping burned areas using these Brazilian satellites with high temporal coverage is useful for conservation units and priority areas in Brazil, aligning with national plans for monitoring the Cerrado. The Action Plan for the Prevention and Control of Deforestation and Wildfires in the Cerrado (PPCerrado), established in 2010 under Law No. 12.817/2009 (Brazil, 2009), which concerns the national policy on climate change, represents a significant milestone in Brazilian environmental policy. Among the strategic contributions of the National Institute for Space Research (INPE) within the framework of the PPCerrado, noteworthy initiatives in line with this research include the development of a biome-scale annual monitoring system of the Cerrado vegetation, covering all its land use and land cover classes, and the implementation of an automated process to estimate areas affected by fires every 15 days, using low-resolution imagery. These efforts underscore a commitment to protecting the environment and mitigating the impacts of deforestation and wildfires in this hotspot for global conservation.

Another INPE task related to PPCerrado is the Near Real-Time Deforestation Detection System (DETER), whose daily production data is forwarded to national enforcement agencies such as IBAMA (Brazilian Institute of Environment and Renewable Natural Resources). DETER methodology is based on photo-interpretation and uses WFI images to generate alerts on deforestation, forest degradation, and wildfire scars due to the high temporal resolution of this sensor. DETER also generates alerts for savanna vegetation, a typical formation of the Cerrado. Before 2015, DETER relied on MODIS imagery, which produced alerts with a minimum mapping unit (MMU) of 25 ha. Since then, WFI has become the primary data source and now provides a more refined MMU of 3 ha (Almeida et al., 2022).

While this study represents an early investigation of a specific area within the Cerrado, it underscores the critical importance of advancing research that employs WFI-based methods aboard the CBERS-4, CBERS-4A, and AMAZONIA-1 satellites to strengthen Brazilian sovereignty and environmental protection. The main objective of this article is to present a dataset of WFI imagery and demonstrate its use by applying the RF algorithm to the BGR NIR bands. While this is a preliminary approach using such data, it serves as a basis for ongoing refinement and provides valuable insight into the capabilities of WFI data to generate an automatic method for mapping burned areas. In this paper, a multi-temporal dataset based on a time series consisting of multispectral bands and spectral indices aggregated to a regular grid with cells of 500 m x 500 m. The time span is three years, from 2020 to 2022, and the study area is the Chapada dos Veadeiros National Park (CVNP) in the state of Goiás, Brazil.

## 2. Data and methods

### 2.1 Chapada dos Veadeiros National Park

The Chapada dos Veadeiros National Park (CVNP) (Figure 1), established in 1961 by Brazilian Federal Decree No. 48.875, is located in the State of Goiás. An important characteristic of the CVNP is its function of supplying several aquifers and tributaries located in the highest portion of the Central Plateau in the Tocantins Araguaia Hydrographic Region, which integrates the Upper Tocantins basin. The CVNP has been the target of expansions and reductions throughout its history, and its current configuration, with 240,611 ha, was defined in 2017 by the Decree of June 5. The body responsible for its management is the Chico Mendes Institute for Biodiversity Conservation (ICMBIO), linked to the Ministry of Environment and Climate Change (MMA). The main phytophysiognomies inside the park are related to savanna formations, but the official Management Plan (ICMBIO, 2021) clarifies that there is a diverse mosaic of vegetation present in the area.

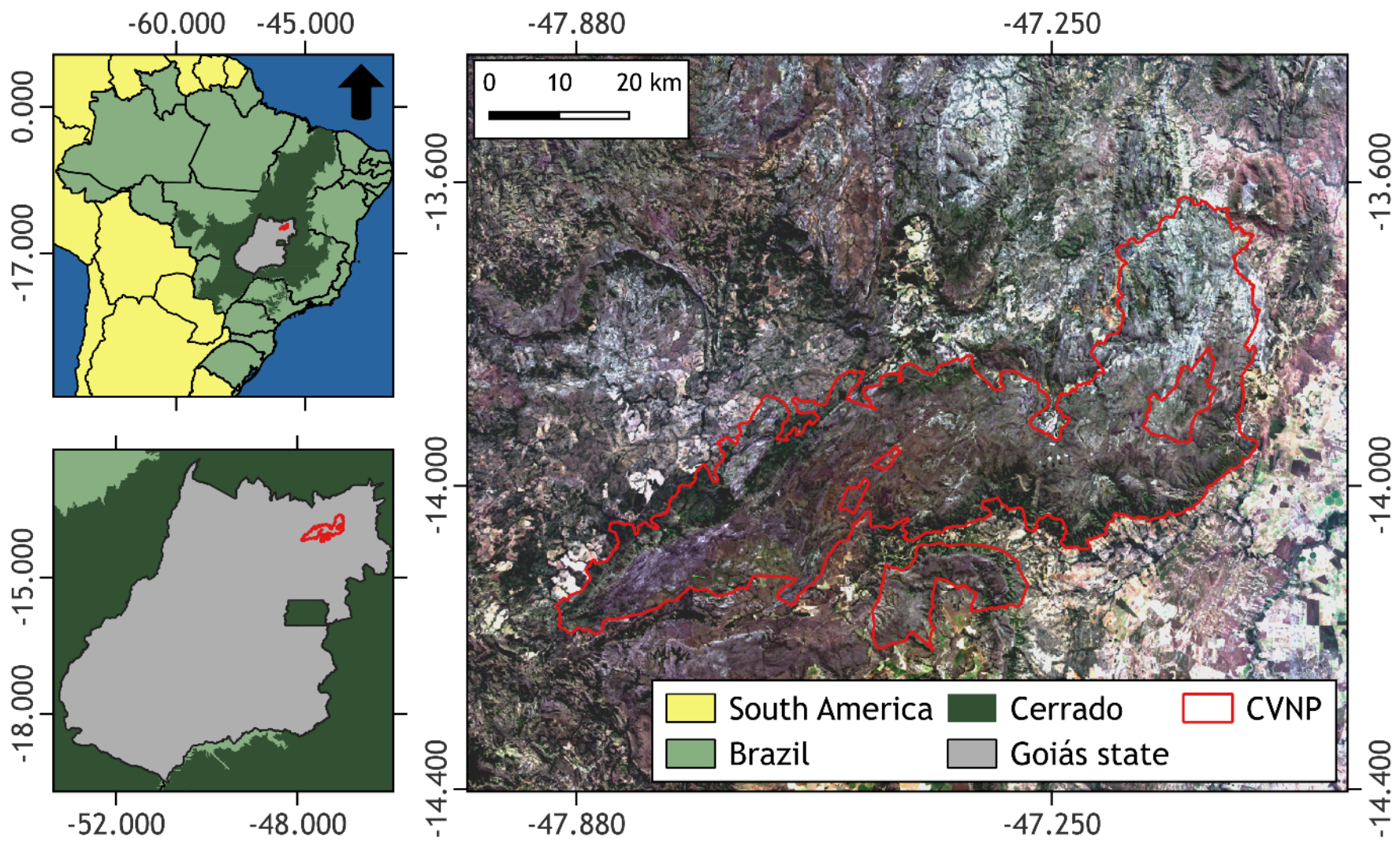

**Figure 1.** Location of the CVNP in the northeastern part of the State of Goiás.

## 2.2 Methodology

The methodology for creating our dataset can be roughly outlined in two stages (raster and vector processing), as shown in Figure 2. To evaluate the characteristics of the dataset, in a third step, different types of labeled data were tested with the RF algorithm, available in the Scikit library (Pedregosa et al., 2011) for Python.

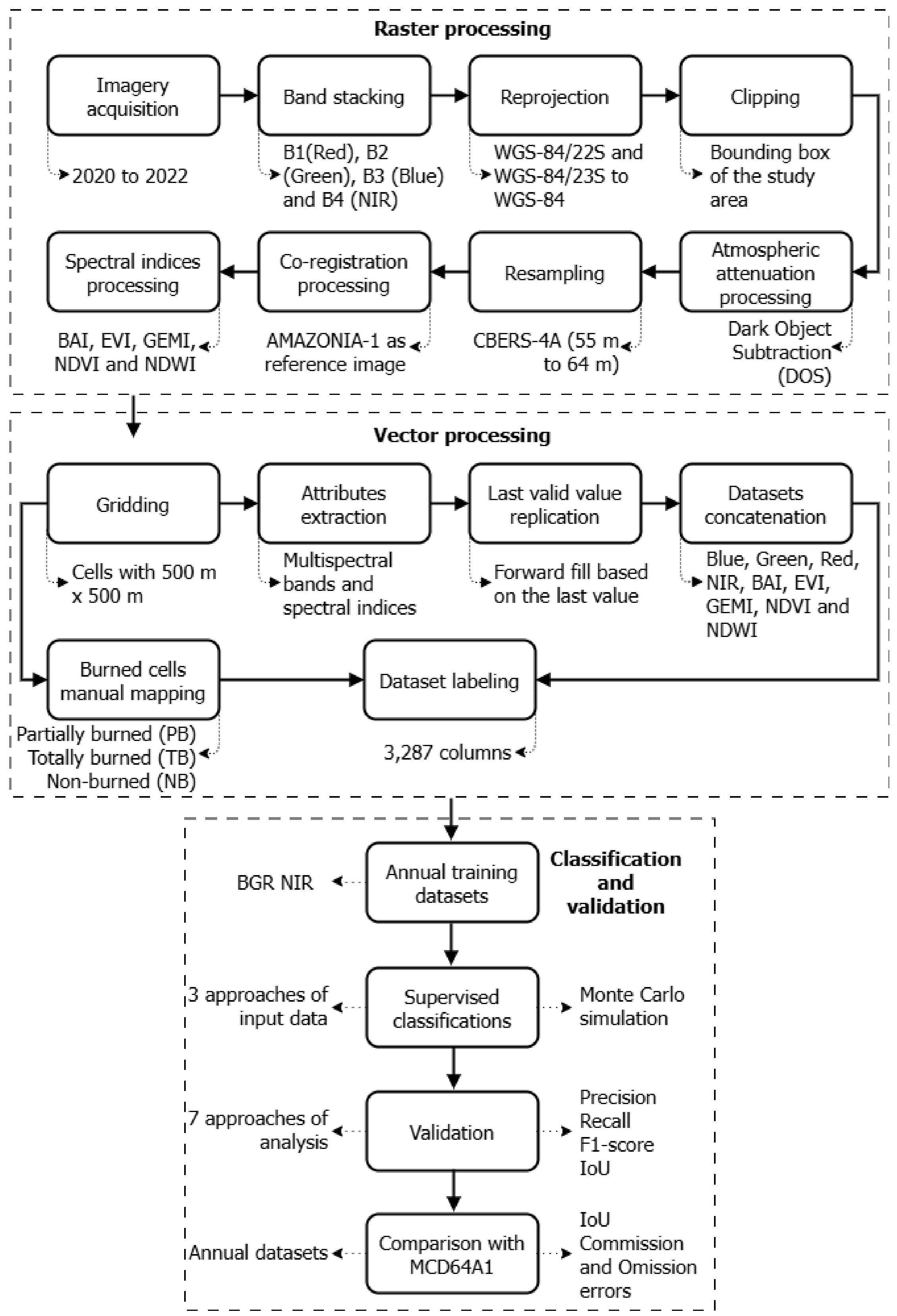

**Figure 2.** General flowchart of the methodology.

### 2.2.1 Imagery acquisition

We downloaded images from the INPE catalog using the Cbers4asat package (Russo, 2023) for serial image downloads based on API query parameters. We set the image acquisition period from January 1st, 2020, to December 31st, 2022, and downloaded all images (BGR NIR bands) from the WFI sensor on board CBERS-4, CBERS-4A, and AMAZONIA-1 satellites at Level-4 (L4), which designates orthorectified products (Oldoni et al., 2022). We set the cloud coverage to 100% to reflect the general scene cloud coverage, which is not necessarily the same as the coverage in our study area. The study area has eight paths observed by the CBERS-4A swath at row 132, and eight paths observed by CBERS-4 at row 117. For AMAZONIA-1, only one path covers the area at row 17. Based on these criteria, we downloaded 382 images, of which 147 were discarded due to clouds, insufficient swath coverage, and repeated images from the same day, resulting in 235 images. In 2020, only CBERS-4A images were included; in 2021, both CBERS-4A and AMAZONIA-1 were included; and in 2022, images from all three satellites were included as the INPE catalog started to include CBERS-4 images that year.

### 2.2.2 Band stacking

INPE provides the blue, green, red and NIR bands for download, and the analyst has to merge them, if necessary, into a stacked raster with the desired bands. We merged the four bands of all images using the Rasterio library for standardizing the sequence as RGB NIR.

### 2.2.3 Reprojection

We reprojected the stacked images from WGS-84/22S (EPSG: 32722) and WGS-84/23S (EPSG: 32723) to WGS-84 (EPSG: 4326) using the nearest neighbor resampling method, which ensures the preservation of the original digital number (DN) values.

### 2.2.4 Clipping

The coordinates of the bounding box were defined based on a buffer zone of 10 km from the boundaries of the CVPN. The buffer zone was established in accordance with Brazilian legislation, which recommends a buffer zone of up to 10 km around conservation units (Brazil, 2002a; Brazil, 2002b). After the data reprojection, we clipped all the images to the same bounding box to optimize further processing. In this step, we used the Os, Rasterio and Shapely libraries for Python.

### 2.2.5 Atmospheric attenuation processing

We used Python to implement an atmospheric attenuation method based on DOS (Dark Object Subtraction), as described by Chavez-Jr (1988) and Nunes et al. (2019). Table 1 lists the technical specifications by band of the WFI sensors used in a first radiometric calibration step, which aims to convert the data from DN to radiance. We needed to estimate the gain and offset factors for the isolated multispectral bands of each image applying Equations 1 and 2.

**Table 1.** Technical specifications of the WFI.

| Band | Wavelength ($\mu$) | Maximum radiance (LMax) | Minimum radiance (LMin) | Maximum Reflectance (ρMax) |
|---|---|---|---|---|
| Blue | 0.45-0.52 | 343.4 | 35.3 | 0.6 |
| Green | 0.52-0.59 | 361.2 | 25.7 | 0.7 |
| Red | 0.63-0.69 | 306.9 | 12.9 | 0.7 |
| NIR | 0.77-0.89 | 243.4 | 8.9 | 0.8 |

$$G = \frac{LMax - LMin}{DNMax - DNMin} \quad (1)$$

$$O = LMin - G * DNMin \quad (2)$$

*G:* band gain factor; *O:* band offset factor; *DNMax* and *DNMin* correspond to the maximum and minimum pixel values of each input band.

We applied the linear function described by Equation 3 to convert the original band data from DN to top-of-atmosphere (TOA) radiance.

$$Lb = G * DN + O \quad (3)$$

*Lb*: pixel radiance value output; *DN*: each input pixel digital number.

We used the Ephem (Rhodes, 2011) library in Python to estimate the Earth-Sun distance for each image, according to metadata included in the strings from the image title, corresponding to the acquisition date information. At this point, we set the observer's latitude and longitude to the centroid of our study area and standardized the image acquisition time to 13:55:00 UTC. We implemented Equation 4 to estimate the solar irradiance. In sequence, Equation 5 was applied to calculate the dark object radiance required by the DOS method.

$$ESUN = \pi * d^2 * \frac{LMax}{\rho Max} \quad (4)$$

*ESUN:* extraterrestrial solar irradiance; $d^2$: Earth-Sun distance.

$$L1\% = \frac{0.01 * ESUN * cos(\theta)}{\pi * d^2} \quad (5)$$

*L1%:* dark object radiance; $\theta$: solar zenith angle.

We performed the subtraction between the generated rasters containing the dark object radiance and the images containing the TOA radiance to attenuate atmospheric noises. We applied Equation 6 and finally the results were converted to surface reflectance using Equation 7.

$$Lp = Lb - L1\% \quad (6)$$

*Lp*: subtraction radiance.

$$\rho = \frac{\pi * (Lb - Lp) * d^2}{ESUN * cos(\theta)} \quad (7)$$

$\rho$: surface reflectance.

**2.2.6 Resampling**

All CBERS-4A images were resampled from 55 m to 64 m in order to equalize the spatial resolution of the three satellites because it allows us to retain the CBERS-4 and AMAZONIA-1 data without resampling. We used the nearest neighbor as the resampling method. For each resampled image, we set the grid from the WFI on board AMAZONIA-1 as reference to fit the generated data with 64 m from CBERS-4A satellite. We also performed the grid alignment between CBERS-4 (target) and AMAZONIA-1 (reference) data.

**2.2.7 Co-registration processing**

We employed the Python library Arosics (Automated and Robust Open-Source Image Co-Registration Software) (Scheffler, 2017) for co-registration. We set the "grid resolution" parameter to 30 pixels and the "window size" parameter to 400 x 400 pixels. By using these parameters, a total of 2,772 CPs were plotted, but not necessarily all of them were used, since the Arosics algorithm disregards possible outliers or points of non-correspondence.

**2.2.8 Spectral indices processing**

We choose to process the spectral indices shown in Table 2: Normalized Difference Vegetation Index (NDVI), Global Environmental Monitoring Index (GEMI), Normalized Difference Water Index (NDWI), Enhancement Vegetation Index (EVI) and the Burned Area Index (BAI), all of them derived from visible and NIR bands.

**Table 2.** Spectral indices applied to the WFI bands and used for constructing the dataset.

| Spectral index | Equation | Reference |
| --- | --- | --- |
| NDVI | $\frac{NIR - Red}{NIR + Red}$ | Rouse et al. (1974) |
| GEMI | $\gamma = \frac{2(NIR^2 - Red^2) + (1.5 * NIR) + (0.5 * Red)}{NIR + Red + 0.5}$ <br> $GEMI = \gamma * (1 - 0.25) - \frac{Red - 0.125}{1 - Red}$ | Pinty & Verstraete (1992) |
| NDWI | $\frac{Green - NIR}{Green + NIR}$ | Mcfeeters (1996) |
| EVI | $2.5 * \frac{NIR - Red}{(NIR + C_1 * Red - C_2 * Blue + L)}$ <br> $C1 = 6, C2 = 7.5, L = 1$ | Huete et al. (2002) |
| BAI | $\frac{1}{(0.1 - Red)^2 + (0.06 - NIR)^2}$ | Chuvieco et al. (2002) |

The indices adopted try to cover specific burned area indices estimated from the visible and NIR bands (in this case, the BAI proposed by Chuvieco et al. (2002)), a simpler vegetation index (NDVI) and an enhanced vegetation index (EVI), a global index (GEMI) and an index taking into account the normalized difference between the green and NIR bands (in this case the NDWI was used, since there are similarities between the spectral behavior of some water bodies and the burned areas in these two bands). According to Duan et al. (2024), the use of multitemporal data can help reduce potential confusions between burned areas on vegetation cover and water, as it captures phenological information over time. Despite these predefined indices, the dataset can also be used to generate and explore other indices using the RGB and NIR bands, as these spectral data are available.

### 2.2.9 Gridding

We adopted the MODIS spatial resolution as the MMU to generate a vector grid with 500 m x 500 m cells size. The generated regular grid has 239 columns x 163 rows, which gives a total of 39,957 cells.

### 2.2.10 Attributes extraction

We used the Os, Geopandas (Jordahl et al., 2020) and Rasterstats libraries in Python to extract the mean zonal statistics from the pixels within the cells, taking into account 235 images from 2020 to 2022. We adopted the blue, green, red and NIR bands, as well the BAI, EVI, GEMI, NDVI and NDWI spectral indices as references for data integration.

### 2.2.11 Data replication

All the 235 valid observations were integrated over the period from January 1, 2020 to December 31, 2022. As mentioned earlier, the number of images available at the INPE catalog is different for each year, and therefore, after extracting the attributes, we obtained datasets that can be understood as irregular time series, because the observations are not equally spaced in time. Burned areas are understood as abrupt and non-permanent changes on the Earth's surface, and among the available interpolation techniques applied to RS data, we choose a forward fill replication because these targets do not necessarily occur between two observations. So, every gap between two valid observations was filled with the value of the first observation, as illustrated in Figure 3.

a)

| Image sequence | 1 | 2 | 3 | 4 | … |
|---|---|---|---|---|---|
| Date of observation | Month x day 1 | Month x day 3 | Month x day 6 | Month x day 7 | … |
| Observed value | a | b | c | d | … |

b)

| Date sequence | Month x day 1 | Month x day 2 | Month x day 3 | Month x day 4 | Month x day 5 | Month x day 6 | Month x day 7 | … |
|---|---|---|---|---|---|---|---|---|
| Replicated value | a | a | b | b | b | c | d | … |

**Figure 3.** An illustration of the adopted replication method. a) Before replication (4 images and 4 columns); b) after replication (4 images and 7 columns).

The methodology addresses a dataset with temporal irregularity, where the pattern of irregularity varies across years due to differing image availability. Each observation is replicated to ensure uniformity across datasets, enabling the Random Forest (RF) algorithm to operate with a consistent set of temporal observations. The primary goal of this choice is to assess the algorithm's performance using the RGB NIR spectral data from the same year. Specifically, the years 2020, 2021, and 2022 are analyzed independently. Models constructed from data of one year are not generalized to other years, and no unified dataset containing data from all years is pre-compiled to create a specific model for post-application to each year separately. Given that burned areas in the study area mainly occur between July and September (Mataveli et al., 2018), future studies using RF could consider semester-based approaches rather than annual ones. Additionally, testing other algorithms may help address the temporal irregularity of this dataset and the incidence of burned areas.

### 2.2.12 Datasets concatenation

At this stage, after replicating the data for each spectral layer, 9 ARD (Analysis Ready Data) sets comprising 1,096 columns were generated, with each column representing one day. To standardize the observations per dataset, we decided to omit February 29th from 2020 because it is a leap year. We then concatenate the datasets in the order shown in Table 3.

**Table 3.** Sequence of concatenated data and object indices.

| Data | Temporal range | Start of the column title | Column indices |
|---|---|---|---|
| Blue | Jan 1st 2020 - Dec 31st 2022 | a_... | [2:1096] |
| Green | Jan 1st 2020 - Dec 31st 2022 | b_... | [1097:2191] |
| Red | Jan 1st 2020 - Dec 31st 2022 | c_... | [2192:3286] |
| NIR | Jan 1st 2020 - Dec 31st 2022 | d_... | [3287:4381] |
| BAI | Jan 1st 2020 - Dec 31st 2022 | e_... | [4382:5476] |
| EVI | Jan 1st 2020 - Dec 31st 2022 | f_... | [5477:6571] |
| GEMI | Jan 1st 2020 - Dec 31st 2022 | g_... | [6572:7666] |
| NDVI | Jan 1st 2020 - Dec 31st 2022 | h_... | [7667:8761] |
| NDWI | Jan 1st 2020 - Dec 31st 2022 | i_... | [8762:9856] |

The first column is the "id" information of each instance.

### 2.2.13 Burned cells manual mapping

Considering that the MMU refers to the size of the cells in the regular grid, it is possible to understand that not all cells may be totally affected by fire, since cells located on the edges of the main affected area, such as on the fingers and flanks, may be partially affected, as well as non-burned islands and smaller spot fires. Consequently we chose to include a label called partially affected cells, as this can affect the supervised classification process and analysis.

We manually collected burned areas within the cells identified by image interpretation for all years using QGIS 3.22.12. We labeled the cells in "totally burned" (TB), "partially burned" (TB) and “non-burned” (NB) classes. The visual image analysis was conducted using the R(R)NIR(G)B(B) color composition to enhance burned areas, which are highlighted as dark areas, as illustrated in Figure 4.

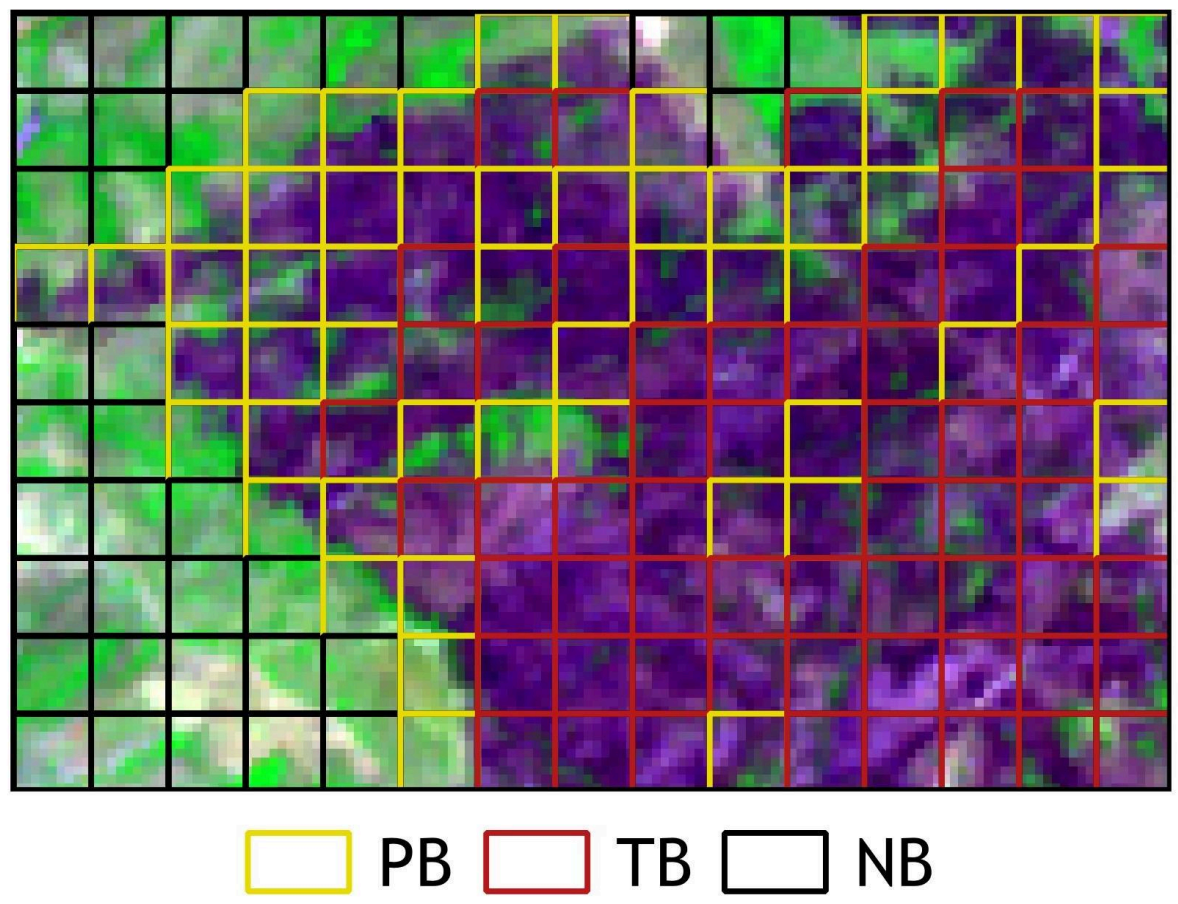

**Figure 4.** Examples of partially burned (PB), totally burned (TB) and non-burned (NB) cells manually classified.

### 2.2.14 Dataset labeling

Given the mapping with the categorical labels PB, TB or NB assigned to the cells with the respective day when the area inside each cell was burned, we used the Pandas library to concatenate these columns to the dataset. Table 4 illustrates the index of the columns titles and their description.

**Table 4.** Description of the label columns.

| Column | Description | Column index | Type |
|---|---|---|---|
| date_20 | Burning date in 2020 (%d-%m-%y) | [9856] | String |
| date_21 | Burning date in 2021 (%d-%m-%y) | [9857] | String |
| date_22 | Burning date in 2022 (%d-%m-%y) | [9858] | String |
| 20_labels | Labels TB, PB or NB for the year 2020 | [9859] | String |
| 21_labels | Labels TB, PB or NB for the year 2021 | [9860] | String |
| 22_labels | Labels TB, PB or NB for the year 2022 | [9861] | String |

Date format: %d-%m-%y

### 2.2.15 Annual training datasets

We adopted the RF supervised classifier (script available in the repository of ScienceDB) to investigate the potential for classifications using annual ARD time series. This involved concatenating the four multispectral bands in 3 different annual datasets (each one containing 1,460 columns), for 2020, 2021 and 2022, which were labeled *a posteriori* in accordance to their temporally equivalent burned cells using different approaches for classifying and validating the results.

### 2.2.16 Supervised classifications

We used GridSearchCV to identify the optimal hyperparameters for our model, focusing on the number of estimators (n_estimators) and the maximum depth (max_depth). The number of decision trees varied between 100, 200, and 300, while the maximum depth was tested at values of 3, 5, and 7. We defined three approaches to investigate the performance of the RF classifications on the annual datasets, as shown in Table 5. Each classification approach was treated as a different type of input data, labeled by combinations of TB, PB, and NB cells. The input sample sizes were based on the least representative class, and all tests used balanced training samples.

**Table 5.** Classification approaches for understanding the performance of the RF classifier on the annual BGR NIR datasets.

| Classification approach | Relabeling | Training set size per class (70%) |
|---|---|---|
| 1st | TB as BA class; PB and NB as NB class | 2020: 1,505<br>2021: 1,726<br>2022: 1,457 |
| 2nd | TB and PB as BA class; NB as NB class | 2020: 3,254<br>2021: 3,432<br>2022: 3,659 |
| 3rd | TB, PB and NB as three different classes | 2020: 1505<br>2021: 1705<br>2022: 1,457 |

For each classification approach, 200 Monte Carlo simulations were performed to provide more stable results for validation. Based on the balanced training set size, we set different random samples in each simulation to capture the variation of the most representative class, despite the subsequent random process of splitting the set into training (70%) and testing (30%) subsets.

### 2.2.17 Validation

Using 30% of the samples for testing and validation, we computed precision, recall, and F1-score metrics for each simulation, using the mean and standard deviation to analyze the results. To assess classification consistency on the same annual dataset, we used the Intersection over Union (IoU) metric. Table 6 outlines the approaches used to validate the classifications. Each validation approach was treated as a different combination of input data for classification and post-classification relabeling.

**Table 6.** Validation approaches for understanding the performance of the RF classifier on the annual BGR NIR datasets.

| Validation approach | Classification approach | Validated |
|---|---|---|
| 1st | 1st | TB as BA class; PB and NB as NB class |
| 2nd | 1st | TB and PB as BA class; NB as NB class |
| 3rd | 2nd | TB as BA class; PB and NB as NB class |
| 4rd | 2nd | TB and PB as BA class; NB as NB class |
| 5th | 3nd | TB reclassified as BA; PB and NB as NB |
| 6th | 3nd | TB and PB reclassified as BA; NB as NB |
| 7th | 3nd | TB, PB and NB as three different classes |

Validation approaches 5th and 6th were evaluated by contrasting the reclassification of the three classes into both BA and NB classes, and in the 7th we considered the three original labels as three different classes. All other validation approaches refer to binary products.

### 2.2.18 Validation with external data

We acquired MCD64A1 monthly products and generated a single raster per year containing all burned pixels observed in 2020, 2021, and 2022. For the comparison with the annual MCD64A1 products, the first analysis used two reference datasets per year: one where the BA class is represented by TB cells, and another where both TB and PB represent the BA class. We then randomly selected one of the 200 output generalizations per year from the 5th validation approach (since it returned the best results) and used it to analyze agreements with the MCD64A1 product. We adopted the IoU, commission errors, and omission errors as validation metrics to assess the results.

## 3. Results and discussion

### 3.1 Description of the time series

Oldoni (2022) investigates harmonization methods for WFI imagery from different satellites, highlighting that the most significant spectral inconsistencies occur in the blue and NIR bands when comparing AMAZONIA-1 with CBERS-4 and CBERS-4A. In contrast, the red bands show the highest spectral equivalence. Despite these considerations, Oldoni notes that the WFI bands are comparable to those of Landsat-8/OLI, suggesting new possibilities for future applications. Our study aims to assess the accuracy of supervised classification using annual time series data without spectral calibration, as inter-satellite noise may have a minimal impact on the performance of the RF algorithm.

Figure 5 displays a density chart constructed from the WFI inter-satellite time series spanning from 2020 to 2022 for the Chapada dos Veadeiros National Park, providing a descriptive overview of the observations within the dataset.

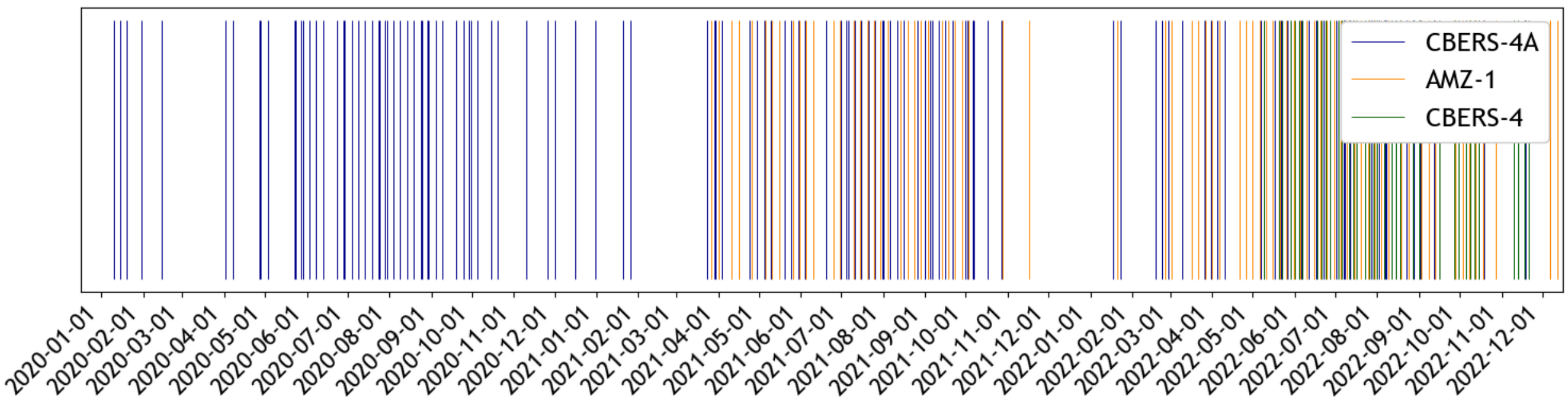

**Figure 5.** Imagery availability displayed as a density chart in the time series.

After removing images with high cloud coverage, swaths covering less than half of the study area, and repeated images,the number of WFI images was reduced from 382 to 235. The amount of data increases over time due to image availability in the reference catalog. The 2020 dataset consists entirely of CBERS-4A images, 2021 includes CBERS-4A and AMAZONIA-1 images, and the INPE catalog recently added CBERS-4 images in 2022, though this satellite has been operational since 2014. Figure 6 shows the distribution of images by satellite, assessing the monthly representativeness of each satellite over the time series.

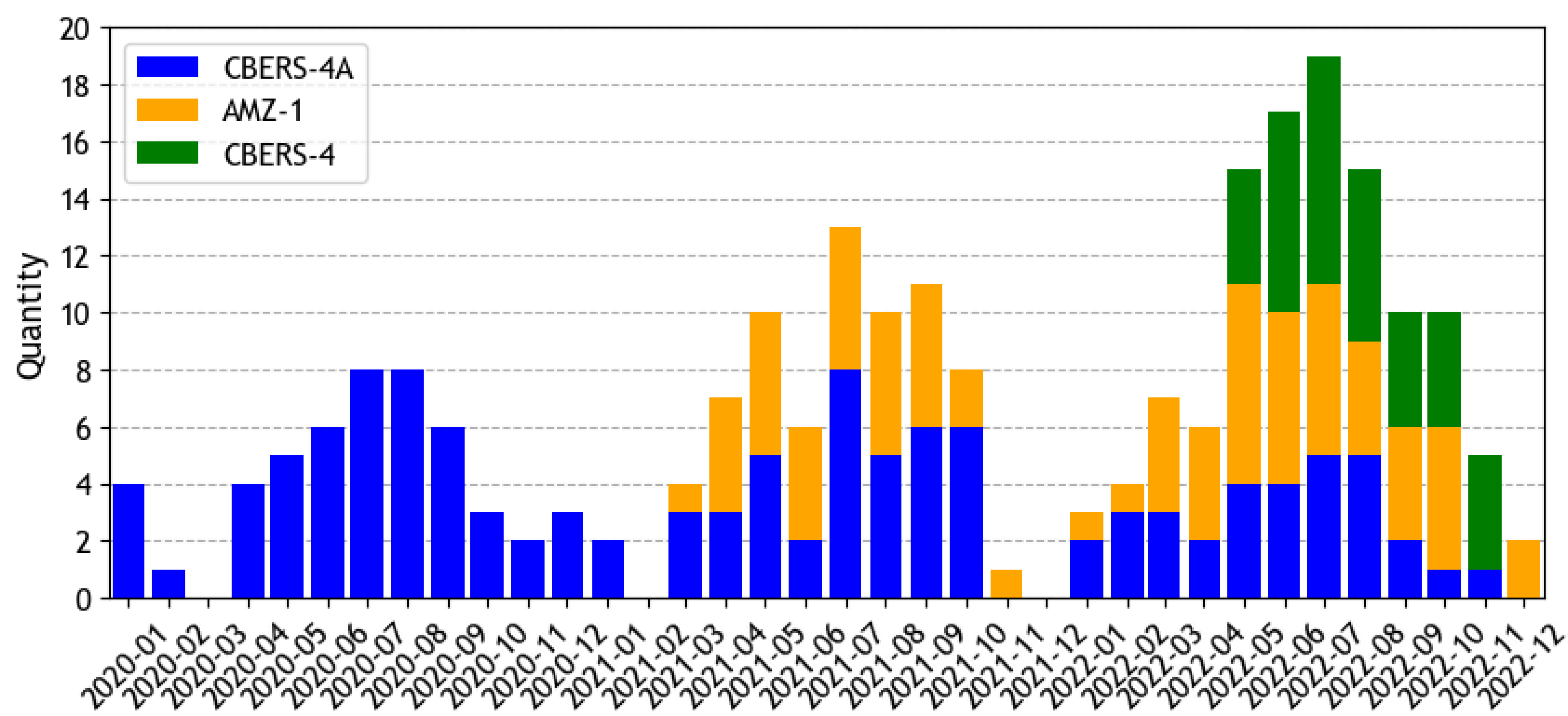

**Figure 6.** Monthly distribution of WFI images per satellite along the time series.

A total of 122 images from CBERS-4A are available in the INPE catalog. Our study area has been covered by AMAZONIA-1 since March 27, 2021, contributing 75 WFI images to the time series, while 38 images from CBERS-4 were included, as they became available on May 9, 2022. The period between November and March records the fewest images due to cloud cover during the rainy season. July stands out with the highest number of images, and April to October accumulates the largest number of images, aligning with the peak fire season in the Brazilian Cerrado and thus serving as the primary window for observing burned areas. Figure 7 clarifies this, showing the monthly count of manually mapped burned cells (TB and PB).

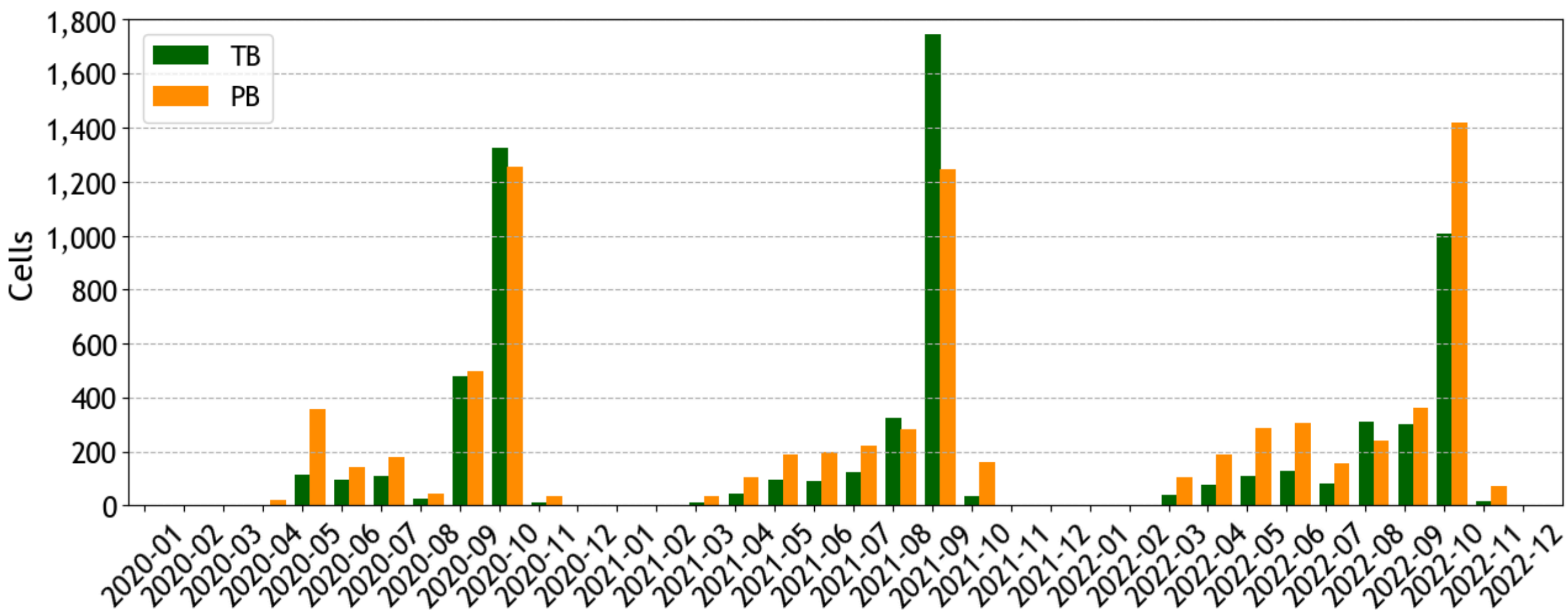

**Figure 7.** Monthly distribution of TB and PB cells along the time series.

Figure 7 reveals gaps in fire observations during the first semesters. In the Cerrado, the natural fire regime occurs mainly during the transition from dry to wet seasons, mainly between May and October, with lightning as the ignition source (Fidelis et al., 2018). However, human-induced fires, despite the possibility of occurring in unfavorable seasons, are not controlled by natural factors such as rain (Ramos-Neto & Pivello, 2000). Mataveli et al. (2018) analyzed the months with the highest concentration of fires during the historical period from 2002 to 2015, identifying May to October as the peak period for fire hotspots across most of the Cerrado. Fidelis et al. (2018) note that in the CVNP region, rains typically begin between late September and early October, although the dry season can extend into mid to late October. This happened in 2017, for example, when a megafire event burned the CVNP.

Figure 7 shows that the years 2020, 2021, and 2022 follow these patterns, as the observed fire peak in 2020 occurred in October, when more than 2,500 cells (625 $km^2$) were totally or partially burned. In 2021, September registered the highest number of these cells, and it was the month with the highest proportion of burned cells among the three years, with nearly 3,000 (750 $km^2$). Similarly to 2020, in 2022 the fire peak was recorded in October, with around 2,400 cells (650 $km^2$). Overall, the other burning events are mainly between April and August, with the lowest number of burned cells occurring from November to March, a period that covers the wet season (spring and summer).

### 3.2 Classification approaches

Figure 8 was elaborated based on 200 Monte Carlo simulations per year, considering the accuracies of the classification approaches using TB as BA, TB + PB as BA and TB, PB and NB as three different labels, and based on random subsets with 30% of all input samples for validation.

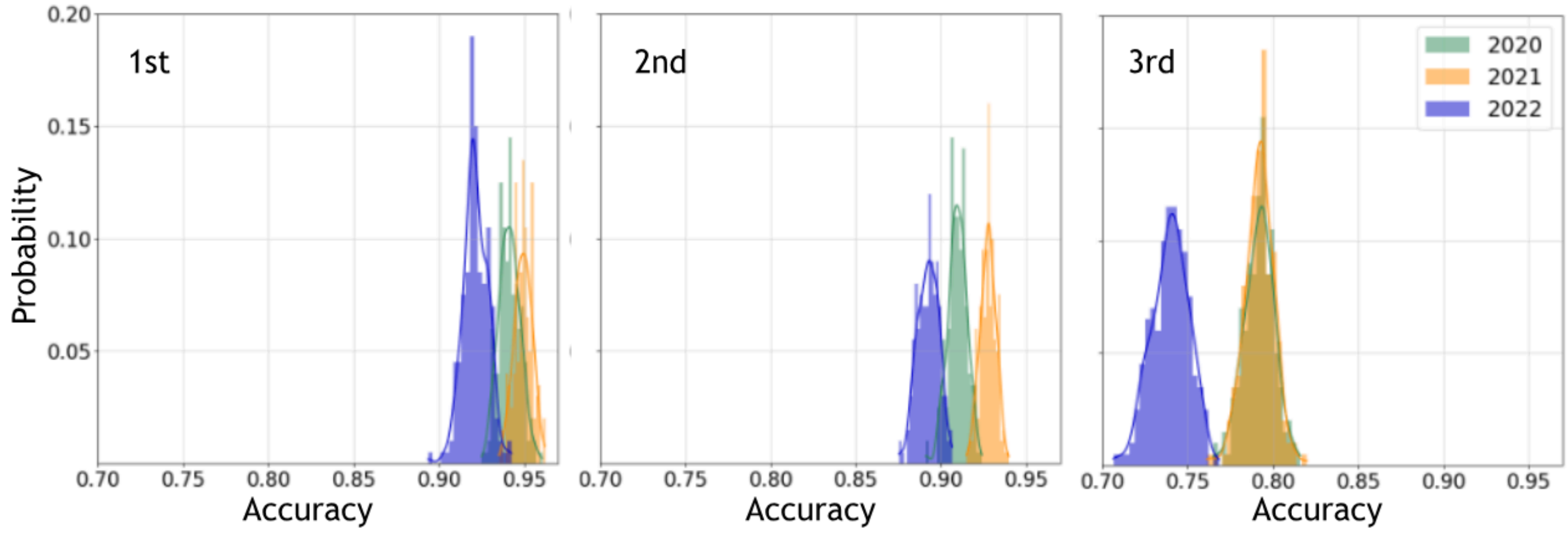

**Figure 8.** Comparison of the probability distributions of the three classification approaches. 1st: TB as BA, 2nd: TB + PB as BA and 3rd: each TB, PB and NB as three different classes.

The first approach (TB as BA class) returned the highest accuracies, with 2022 showing the weakest performance, with a mean accuracy around 0.92 in about 15% of the simulations. The left tail indicates a probability of scoring below 0.91 in less than 5%, while a peak near 0.93 on the right tail suggests a probability of 10%. The intermediate results were obtained using the dataset from 2020, with a mean accuracy of 0.94 in about 10% of simulations, followed by 2021, the year with the highest accuracies in which there is a probability close to 10% to obtain accuracies above 0.95. Across all years, intersection probabilities remain below 1%, converging on an accuracy close to 0.94.

The second approach (TB + PB as BA class) returned lower accuracies compared to the previous analysis, although the probability distributions per year exhibit similarities. In 2022, accuracy ranges from 0.88 to 0.9, with the highest probabilities around 0.89, occurring between 5% to 10%. Accuracies for 2020 are at 0.91, with probabilities exceeding 10%. An intersection occurs between the right tail of 2022 and the left tail of 2020, with probabilities around 3~5% for accuracies near 0.9. Accuracies for 2021 varied from 0.92 to 0.94, with the highest probability at 0.93, around 15%. There is an intersecting region between 2020 and 2021 where probabilities exceed 5% for accuracies between 0.94 and 0.95.

The third approach (TB, PB, and NB as three different classes) exhibited the weakest performances among all approaches. The positions of the probability distributions for 2020 and 2021 differed from previous analyses, showing a large intersection between these years. Both years have accuracies ranging from 0.78 to 0.83, with a probability of obtaining an accuracy of 0.79 in about 15% for 2021 and 10% for 2020. Like the first two approaches, 2022 performed the worst, with no overlap between its right tail and the left tails of 2020 and 2021. This delineates two main behaviors ranging from 0.72 to 0.77 for 2022 and from 0.78 to 0.82 for 2020 and 2021.

We performed the Shapiro-Wilk test with a significance level (alpha) of 5% ($\alpha = 0.05$). The test evaluates whether the data can reasonably be assumed to follow a normal distribution, and all results supported this hypothesis. We then conducted paired Kolmogorov-Smirnov tests to verify whether the annual classifications per approach could be considered statistically similar between them (H0) with $\alpha = 0.05$, and the only similar performances were identified between the years 2020 and 2021 using the third approach. The results of these tests are in line with the qualitative analysis of Figure 8.

Based on the analysis of each approach, we conclude that the classifier performs better when TB cells represent the BA class (the 1st approach), and that the most accurate dataset is 2021. Predicting PB cells is challenging due to higher spectral heterogeneity, resulting in noise from burned area spectral responses caused by adjacent targets within the cells. Alternatively, using TB cells representing the BA class may limit the classifier's ability to detect heterogeneous targets, potentially underestimating burned areas, as PB cells may be classified as NB. This is confirmed by analyzing the precision and recall metrics of the third approach shown in Table 7.

**Table 7.** Mean results estimated based on 200 classifications per annual dataset using 30% for validation per 1st, 2nd and 3rd approaches.

| | 2020 | 2021 | 2022 |
|---|---|---|---|
| **1st** | | | |
| F1-score | 0.94 | 0.95 | 0.92 |
| Precision BA | 0.94 | 0.94 | 0.9 |
| Recall BA | 0.94 | 0.96 | 0.95 |
| Precision NB | 0.94 | 0.96 | 0.95 |
| Recall NB | 0.94 | 0.94 | 0.89 |
| **2nd** | | | |
| F1-score | 0.91 | 0.93 | 0.89 |
| Precision BA | 0.94 | 0.94 | 0.88 |
| Recall BA | 0.87 | 0.91 | 0.9 |
| Precision NB | 0.88 | 0.91 | 0.9 |
| Recall NB | 0.94 | 0.95 | 0.88 |
| **3rd** | | | |
| F1-score | 0.79 | 0.79 | 0.74 |
| Precision NB | 0.84 | 0.87 | 0.87 |
| Recall NB | 0.89 | 0.90 | 0.8 |
| Precision PB | 0.71 | 0.7 | 0.62 |
| Recall PB | 0.69 | 0.7 | 0.66 |
| Precision TB | 0.82 | 0.81 | 0.75 |
| Recall TB | 0.79 | 0.78 | 0.76 |

Figure 9 presents the precision and recall analysis for the third approach. The NB class achieved the highest performance, ranging between 0.8 and 0.9 for both metrics. The TB class closely followed with values between 0.75 and 0.8. However, the PB class, representing an intermediate category, showed lower performance, ranging between 0.7 and 0.6. Specifically for the NB class, a trend emerged where recall outperformed precision in 2020 and 2021, but this shifted in 2022, indicating a decline in the classifier's ability to detect burned areas accurately from new data. Conversely, the TB class initially had precision exceeding recall in 2020 and 2021, but recall became slightly superior in 2022. The PB class showed the lowest performance across all years, with 2021 having a more balanced precision-recall relationship, while 2020 favored precision and 2022 favored recall.

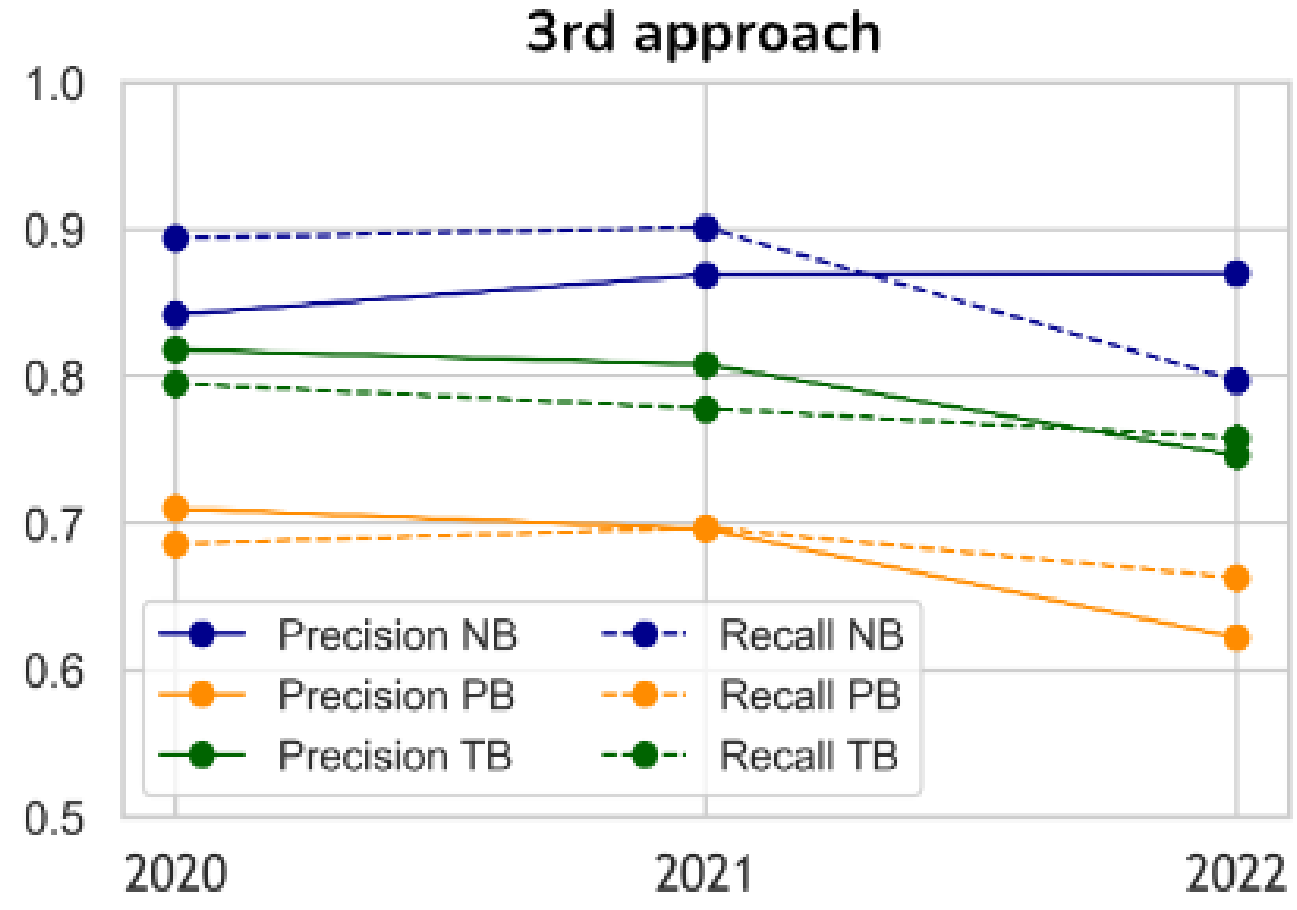

**Figure 9.** Comparison between classes of the third approach.

Figure 10 illustrates the mean precision (Figure 10a) and recall (Figure 10b) for the first and second approaches for both BA and NB classes and for all years. Figure 10a allows us to verify that in 2020 and 2021, the 1st approach shows consistently high precision for both classes, with values ranging from 0.94 to 0.96. However, in 2022, there is a decrease in precision for the BA class, dropping to 0.9, while precision for the NB class remains stable at 0.95. Conversely, the 2nd approach shows slightly lower overall precision compared to the 1st approach. In particular, in 2022, the precision for both the BA and NB classes decreased significantly, with the BA precision dropping to 0.88 and the NB precision dropping to 0.9. This suggests that while the 1st approach maintains higher precision throughout the years, it experiences a decline in 2022, particularly in the BA class. The 2nd approach, while maintaining lower precision scores, also experiences a decline in precision in 2022.

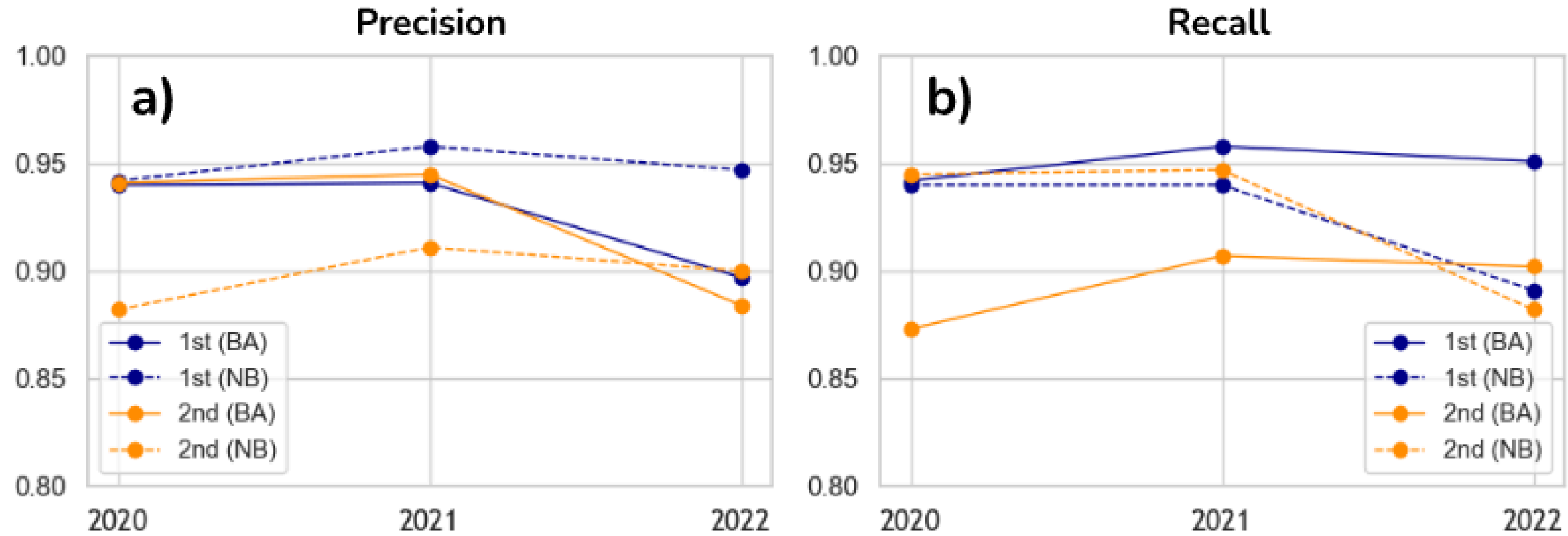

**Figure 10.** Comparison between precision and recall for 1st and 2nd approaches.

The recall analysis from Figure 10b shows that in 2020 and 2021, the 1st approach returned highest recalls, both classes with 0.94 in 2020 and varying from 0.94 (NB) to 0.96 (BA) for 2021. In 2022, although the recall of the BA class remains stable compared to 2021, there is a decrease in the recall of the NB class, making it the year with the lowest recall performance for this class. The 2nd approach yields lower recall compared to the 1st approach, with scores ranging from 0.87 to 0.91 for BA and 0.94 to 0.95 for NB in 2020 and 2021. In 2022, there is a slight improvement in the BA class under the 2nd approach, rising to 0.9, while the NB class decreases to 0.88. These results suggest that while the 1st approach maintains higher recall overall, it experiences a decline in 2022, impacting both classes. In contrast, although the 2nd approach tends to yield lower recall, it exhibits a slight improvement in the BA class in 2022.

The RF algorithm available in the ScienceDB repository performs hyperparameter optimization by exploring values for the number of trees (n_estimators) and the maximum tree depth (max_depth). Specifically, GridSearchCV evaluates n_estimators with values of 100, 200, and 300 trees, and max_depth with values of 3, 5, and 7. While these parameters are automatically optimized during the classification process, we did not investigate which specific configurations are most frequently selected in the final models. Future research could examine this sensitivity in greater detail, potentially incorporating additional parameters such as min_samples_split (the minimum samples needed to split an internal node), min_samples_leaf (the minimum samples needed at a leaf node), and class_weight (weights associated with classes). Notably, adjusting class_weight could help address issues related to the PB class discussed earlier.

### 3.3 Validation approaches

Table 8 presents IoU results obtained by generalizing the models to the full annual datasets. It includes the overall IoU and IoU for the BA and NB classes. Additionally, for the 7th validation approach, it displays the IoU for the TB, PB, and NB classes. Validation approaches 1 and 2 are generalizations of the 1st classification approach, while validation approaches 3 and 4 are generalizations of the 2nd classification approach. Validation Approaches 5, 6, and 7 are generalizations of the 3rd classification approach.

**Table 8.** Mean results of the IoU metric estimated based on 200 generalizations per annual dataset across the seven validation approaches.

| Classification approach | Validation approach | 2020 | 2021 | 2022 |
|---|---|---|---|---|
| 1 | **1st** | | | |
| | IoU | 0.94 | 0.94 | 0.9 |
| | IoU BA | 0.48 | 0.52 | 0.34 |
| | IoU NB | 0.94 | 0.94 | 0.89 |
| 1 | **2nd** | | | |
| | IoU | 0.94 | 0.95 | 0.91 |
| | IoU BA | 0.58 | 0.68 | 0.52 |
| | IoU NB | 0.93 | 0.95 | 0.9 |
| 2 | **3rd** | | | |
| | IoU | 0.91 | 0.90 | 0.83 |
| | IoU BA | 0.52 | 0.39 | 0.24 |
| | IoU NB | 0.9 | 0.89 | 0.82 |
| 2 | **4th** | | | |
| | IoU | 0.94 | 0.95 | 0.89 |
| | IoU BA | 0.66 | 0.69 | 0.54 |
| | IoU NB | 0.94 | 0.94 | 0.88 |
| 3 | **5th** | | | |
| | IoU | 0.98 | 0.98 | 0.95 |
| | IoU BA | 0.66 | 0.71 | 0.49 |
| | IoU NB | 0.97 | 0.98 | 0.95 |
| 3 | **6th** | | | |
| | IoU | 0.9 | 0.91 | 0.82 |
| | IoU BA | 0.53 | 0.56 | 0.41 |
| | IoU NB | 0.89 | 0.9 | 0.79 |
| 3 | **7th** | | | |
| | IoU | 0.85 | 0.85 | 0.76 |
| | IoU TB | 0.75 | 0.75 | 0.66 |
| | IoU PB | 0.76 | 0.73 | 0.73 |
| | IoU NB | 0.99 | 0.99 | 0.96 |

Figure 11 displays a radar plot comparing overall IoU across validation approaches over the years. The 1st classification approach returned high overall IoU scores, with values of 0.94, 0.94, and 0.89 for the respective years, while the 2nd classification approach produced slightly lower, but still commendable, IoU scores of 0.94, 0.95, and 0.90 over the same period. Both methods demonstrated competitive accuracy.

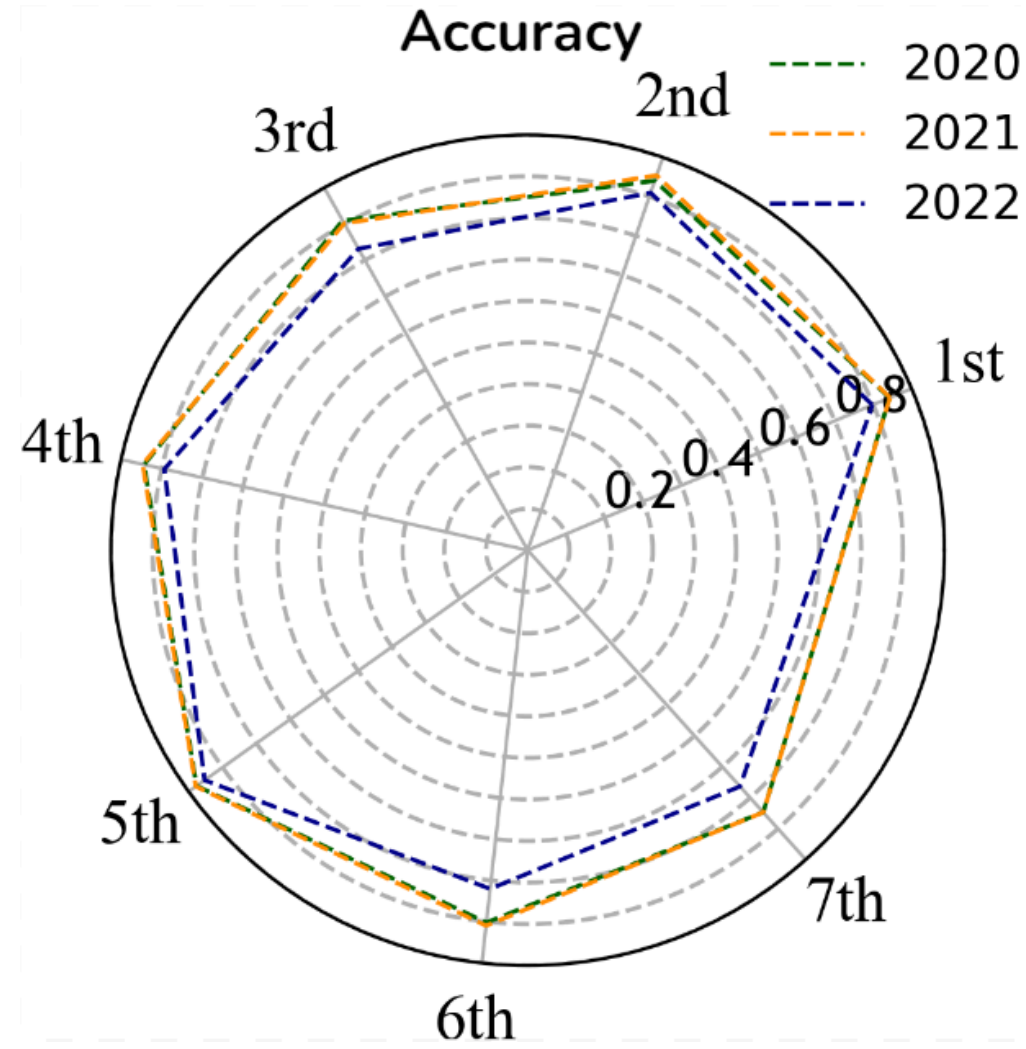

**Figure 11.** Comparison of the IoU results per annual dataset based on the validation approaches scheme shown in Table 4.

The 3rd approach yielded annual IoU scores of 0.90, 0.90, and 0.83, while the 4th approach yielded scores of 0.94, 0.94, and 0.89 for the corresponding years. When evaluated, the 4th approach outperformed the 3rd, with higher IoU scores in all years. In particular, 2021 emerged as the most accurate year for both approaches, with the 4th approach reaching its peak in 2020 and 2021. This analysis highlights the best performance of models trained with TB and PB as BA, applied to test datasets where the BA class includes both TB and PB (4th approach), as opposed to solely TB representing BA.

Concerning the 5th, 6th, and 7th validation approaches, we verify that the 5th approach achieved significantly higher overall IoU scores compared to the others in all three years (0.97, 0.97, and 0.95), highlighting the 5th approach's overall superiority for validation. All approaches performed best in 2021, followed by 2020 and 2022. It should be noted that the performance of the 5th approach was the most stable over the years, including 2022, the year that tends to yield the lowest IoU correspondences.

The evaluation of the overall IoU scores showed that the 5th validation approach is the most efficient. This approach involves generating and generalizing the model based on three classes and when the classification results are then relabeled, with TB representing the BA class and PB and NB representing the NB class. This approach not only improves the accuracy of burned areas classification, but also highlights the importance of nuanced model training and classification refinement. By prioritizing the delineation of specific classes and reclassifying the results, the 5th approach outperformed other approaches. This highlights the importance of tailored model architectures and careful post-processing techniques to optimize classification accuracy for complex tasks such as burned areas mapping.

However, analyzing global IoU can hide important information due to differences in the representativeness of each class. The dataset has 239 columns and 163 rows, which represents 38,957 regular cells. Considering the annual labels TB, PB and NB for 2020 we have, respectively, 2,151 (or 5.52% of the grid), 2,498 (6.41%) and 34,308 (88.06%) cells, for 2021 we have 2,467 (6.33%), 2,437 (6.25%) and 34,053 (87.41%) and for 2022 we have 2,082 (5.34%), 3,146 (8.07%) and 33,729 (86.58%). In other words, in this context of burned areas within regular cells, the most representative class is expected to be NB, covering over 85% of the cells. Figure 12 illustrates the IoU for both classes in approaches 1 to 6 (binary validations), providing insights into the individual performance of each class.

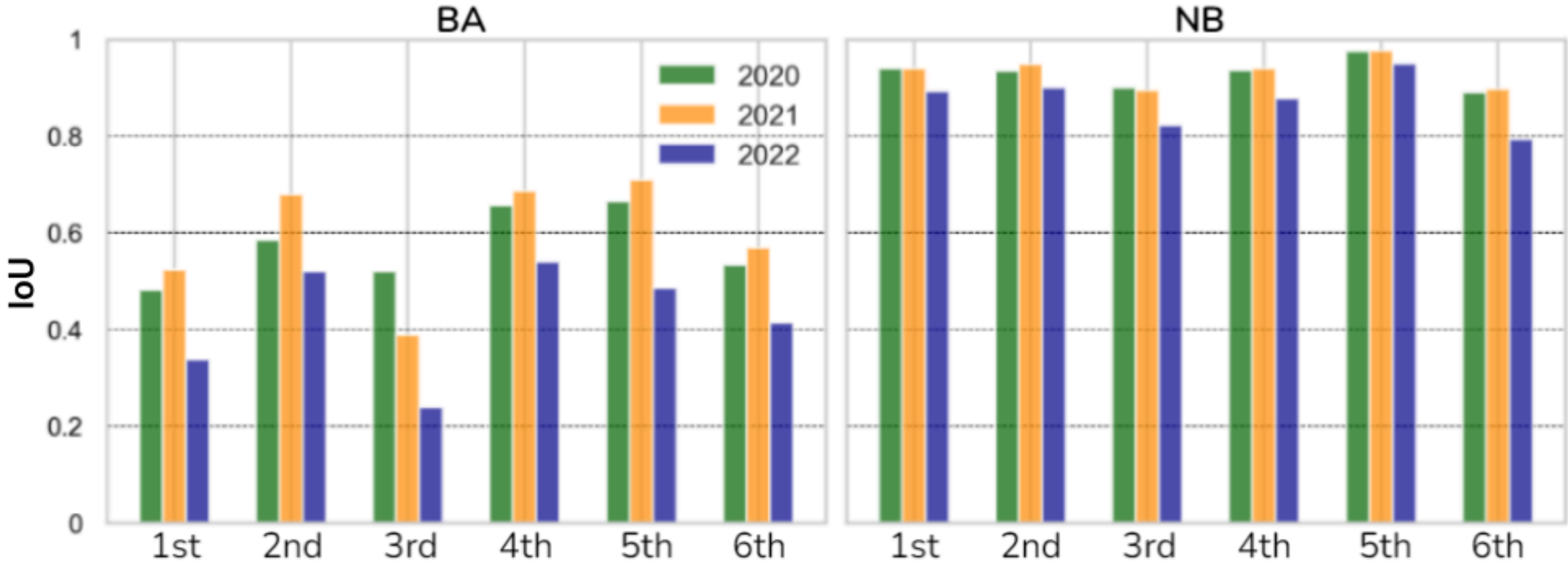

**Figure 12.** Evaluation of the BA and NB classes per binary validation approaches.

First, the IoU scores show that the 2nd approach outperforms the 1st approach in identifying burned areas, as indicated by higher IoU scores for the BA class in all years. However, both approaches show decreased performance in 2022, possibly due to changes in burned areas behavior or data fluctuations. While the 2nd approach excels in BA classification, the 1st approach performs better in NB classification. Comparing the 1st and 2nd approaches, we conclude that it is better to train using TB representing BA and validate by merging TB and PB into the BA class. The analysis of the 3rd and 4th approaches reveals better performance in classifying BA using the 4th approach, where the input data for both classification and validation of this class is represented by TB and PB as BA. The largest difference between the 2nd and 4th approaches appears in 2020, with a 7.2% margin favoring the 4th approach.

Regarding the training with three classes and the 5th (TB reclassified as BA and PB and NB concatenated to NB) and 6th (TB and PB reclassified as BA and NB as NB) validation approaches, the 5th approach gives higher IoU scores for the BA class in all years, with important differences in terms of performance. The largest IoU differences were observed in 2020 (13.2%) and 2021 (14%) in favor of the 5th approach, while this difference was 7.14% in 2022. A similar pattern is seen in the NB class, with higher IoU scores for the 5th approach, reaching 8.57% in 2020, 8.01% in 2021 and 15.63% in 2022. It indicates that the training with three classes is better generalized when, after classification, only the TB label is reclassified as BA, keeping PB and NB as NB. In terms of overall IoU for the BA class among all analyzed approaches, the 4th and 5th validation approaches comparable performances, with the 5th performing better in 2020 and 2021, and the 4th yielding the best performance in 2022, followed by the 2nd approach.

Figure 13 shows the annual IoU scores for the 7th validation approach, which was trained and validated using three classes.

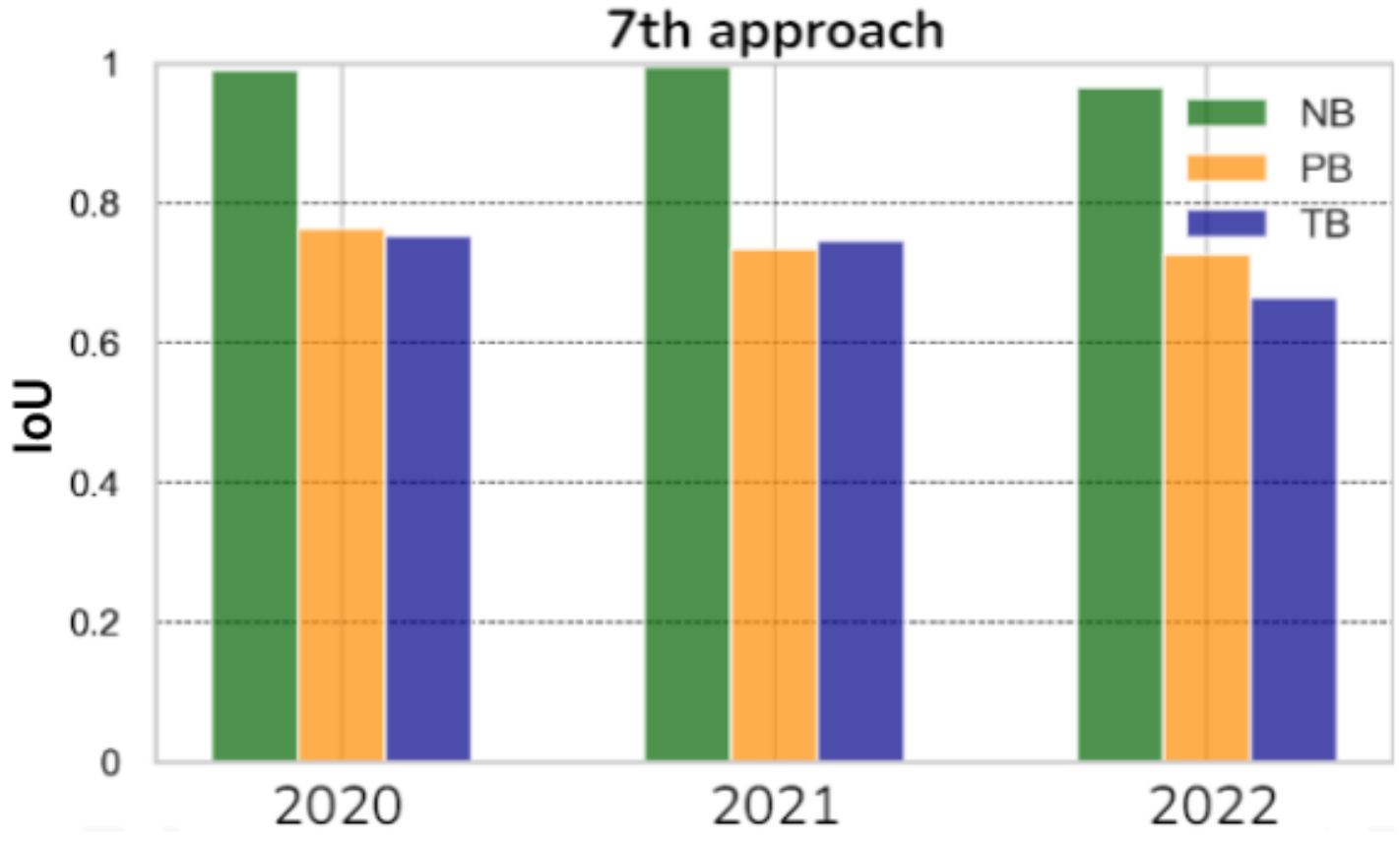

**Figure 13.** Evaluation of the annual results from the 7th approach.

The year 2022 had the lowest values for all TB, PB, and NB classes (0.66, 0.72 and 0.96, respectively). The PB class performed better in 2020 (0.76), and that year also had the highest IoU for the TB class (0.75), with comparable results for both classes. In 2021, TB slightly outperformed PB (0.74 and 0.73), and the largest discrepancy occurred in 2022 (0.72 for PB and 0.66 for TB). For the NB class, IoU values exceeded 0.96 in all years. In conclusion, while NB maintains high performance, variations in the PB and TB labels across years highlight the challenge when facing spectral mixing in regular cells, especially when dealing with PB, which indicates non-homogeneous targets due to the different spatial distribution of burned areas within the cells. Sometimes these cells are more similar to TB cells, sometimes to NB cells.

Table 9 provides data to support a more detailed discussion of the errors of commission and omission of the generalizations, when compared to the manually drawn reference maps. When analyzing the commission errors (CE) for the TB and PB classes, different annual trends are observed: the TB class shows variability, with an initial value of 0.284 in 2020, decreasing to 0.156 in 2021, before increasing to 0.484 in 2022, the highest value recorded for this metric, again reflecting the highest uncertainties for that year, which includes images from all satellites. In contrast, the PB class has consistently higher CE than the TB class across all years, starting at 0.644 in 2020, increasing slightly to 0.65 in 2021, and peaking at 0.725 in 2022, indicating ongoing challenges in minimizing these errors and potential systematic difficulties in accurately identifying this class without overestimation. These CE indicate that more than 64% of all cells classified as PB are not PB in all years.

**Table 9.** Commission and omission error for the 7th approach results.

| | Commission error (TB) | Omission error (TB) | Commission error (PB) | Omission error (PB) | Commission error (NB) | Omission error (NB) |
|---|---|---|---|---|---|---|
| 2020 | 0.284 | 0.154 | 0.644 | 0.22 | 0.0087 | 0.106 |
| 2021 | 0.156 | 0.187 | 0.65 | 0.206 | 0.0076 | 0.095 |
| 2022 | 0.484 | 0.151 | 0.725 | 0.252 | 0.0096 | 0.208 |

For the TB class, omission errors (OE) remain relatively stable over the years, starting at 0.154 in 2020, increasing slightly to 0.187 in 2021, and then decreasing to 0.151 in 2022. These relatively low and consistent values suggest that the generalization performs reliably in minimizing the omission of true positives for the TB class. In other words, about 15% of all TB cells were not detected by the generalizations. The PB class, however, shows a higher variability in the OE, although it is lower compared to the CE. Starting from 0.22 in 2020, the OE decreases slightly to 0.206 in 2021, before increasing to 0.252 in 2022. There is a stable trend to reach OE between 20% and 25%, indicating that at least 1/5 of all PB cells are not correctly classified as PB. An overview of Table 9 shows that the annual trends for both TB and NB cells are to achieve lower omission errors than commission.

The only class that produced more OE than CE in all years is NB. The error percentages for this class are relatively lower due to the greater presence of these cells in the annual datasets, but it can be seen that in 2020 and 2021 the OE were around 10%, while in 2022 this value is doubled. Although burned areas typically occur on cover classes associated with vegetation, there are also possible differences in the nature of that cover (e.g., forest and agricultural land) that may affect the classification of partially burned cells more than completely burned cells. Nevertheless, it is to be expected that even for completely burned cells, there will be different responses from supposedly homogeneous targets (e.g., burned areas) because, as Chuvieco (2019) explains, the evaluation of burned areas by RS is not strictly binary (burned/non-burned), as there are different degrees of burning and different periodicity of image acquisition. Water and shadows may cause spectral confusion in the visible and NIR bands, but time series could minimize these effects (Duan et al., 2024).

Table 10 shows the confusion matrices for all annual generalizations using the 7th approach. In all years, TB cells are more confused with PB than with NB, as indicated by higher false positives represented by PB (296 in 2020, 434 in 2021, and 295 in 2022). Most false positives from PB cells are attributed to TB, suggesting similarities between the two classes. However, the larger number of NB cells makes it the main source of confusion, despite its errors being lower than TB and PB, as shown in Table 9. In 2020, 404 NB cells were misclassified as TB, and 3230 as PB. In 2021, 73 NB cells were misclassified as TB and 3162 as PB. The highest confusion occurred in 2022, with 1108 NB cells misclassified as TB and 5916 as PB. These results indicate that NB is highly confused with PB, contributing to higher CE in PB than in TB, as cells that should be classified as NB are majority detected as PB cells when the models are generalized.

**Table 10.** Confusion matrices for the 7th approach results.

| | 2020 | | | 2021 | | | 2022 | | |
|---|---|---|---|---|---|---|---|---|---|
| | Prediction | | | | | | | | |
| | TB | PB | NB | TB | PB | NB | TB | PB | NB |
| **TB** | 1819 | 296 | 36 | 2005 | 434 | 28 | 1767 | 295 | 20 |
| **PB** | 317 | 1948 | 233 | 297 | 1933 | 207 | 553 | 2353 | 240 |
| **NB** | 404 | 3230 | 30674 | 73 | 3162 | 30818 | 1108 | 5916 | 26705 |

A valuable direction for future studies could be to investigate the influence of land use and land cover classes on the confusion between NB and PB. The statement that time series analysis can cover spectral similarities between targets by incorporating temporal variations may be more applicable to homogeneous targets, such as totally burned and unburned cells. However, the PB class consists of cells with a spectral mixture of burned and unburned areas, where the unburned mixture may represent different Cerrado phytophisiognomies, water, exposed soil, or other classes. Therefore, a key application of the dataset could be its use in a comparative analysis with land use and land cover products, which may provide further insight into why NB cells are confused with PB cells. Alternatively, spectral indices from the dataset could help to assess whether they reduce the confusion between NB and PB when used as classification attributes, as the analysis of the preset paper was based on the RGB NIR bands.

### 3.4 Comparison with MCD64A1

Table 11 shows the IoU scores from comparing annual reference datasets with the MCD64A1 product. For each year, two reference datasets were analyzed: one with TB labels representing the BA class and another combining TB and PB as the BA class. The table also includes results from comparisons between MCD64A1 products and the annual results from the 5th approach.

**Table 11.** Comparison of the annual reference datasets with the MCD64A1 product.

| Reference labels | Target product | IoU | IoU of BA class | IoU of NB class | Commission error (BA) | Omission error (BA) |
|---|---|---|---|---|---|---|
| TB as BA (2020) | MCD64A1 (2020) | 0.876 | 0.453 | 0.935 | 0.493 | 0.189 |
| TB as BA (2021) | MCD64A1 (2021) | 0.82 | 0.455 | 0.897 | 0.44 | 0.289 |
| TB as BA (2022) | MCD64A1 (2022) | 0.838 | 0.334 | 0.916 | 0.542 | 0.445 |
| TB + PB as BA (2020) | MCD64A1 (2020) | 0.873 | 0.551 | 0.927 | 0.166 | 0.379 |
| TB + PB as BA (2021) | MCD64A1 (2021) | 0.831 | 0.516 | 0.894 | 0.126 | 0.441 |
| TB + PB as BA (2022) | MCD64A1 (2022) | 0.799 | 0.363 | 0.881 | 0.18 | 0.604 |
| MCD64A1 (2020) | 5th approach (2020) | 0.867 | 0.464 | 0.932 | 0.227 | 0.461 |
| MCD64A1 (2021) | 5th approach (2021) | 0.811 | 0.463 | 0.895 | 0.282 | 0.432 |
| MCD64A1 (2022) | 5th approach (2022) | 0.798 | 0.301 | 0.892 | 0.593 | 0.46 |

The data analysis reveals a decline in the overall IoU scores from 2020 to 2022, reflecting a gradual divergence between the MCD64A1 product and reference datasets. In 2020, the overall IoU ranges from 0.873 to 0.876, depending on whether PB labels are included; by 2022, these values drop to 0.799 and 0.838, respectively. We highlight that the samples were manually collected by overlaying the vector grid (500 m x 500 m cell size) on the original WFI images with 64 m x 64 m resolution, which imply more detail, such as the need to include PB cells (Figure 4).

In 2020 and 2022 the overall IoUs tend to be higher when the BA class is represented by TB, with the exception of 2021, when the reference containing TB and PB representing the BA class returned a higher IoU.

In 2020, the BA IoU is highest when PB cells are included in the reference dataset (with nearly a 10% difference), where they represent the BA class alongside TB cells. This suggests that incorporating PB areas improves the detection of fire-affected regions, particularly in the dataset with the fewest CBERS-4A images. This improvement may result from PB cells capturing transitional areas that exhibit characteristics of both burned and unburned regions, such as borders and islands within larger burned areas. In 2021 and 2022, the same trend is observed, but the differences between the two reference datasets are smaller. The IoU scores for 2020 and 2021 are similar, despite 2021 having images from both CBERS-4A and AMAZONIA-1 satellites, resulting in more images. As discussed in the previous analysis of the 2022 dataset, a similar trend was observed, with the lowest scores for both overall and BA IoU.

The analysis of CE shows errors above 45% for all annual datasets labeled with TB as the BA class, while these errors decrease to below 20% when TB and PB labels are compared with the MCD64A1 product. Although CE tends to be lower when TB and PB are combined, OE shows the opposite trend. While TB representing BA results in the highest CE, it also produces the lowest OE compared to TB and PB. OE is lowest (0.189) when only TB represents BA in 2020, followed by 2021 (0.289) and 2022 (0.445). The OE increases to 0.379, 0.441, and 0.604, respectively, when BA is represented by both TB and PB. These results align with those of Boschetti et al. (2019), who estimated commission and omission errors for the MCD64A1 product to be around 40.2% and 72.6% globally, and 35.2% and 60% for tropical savannas like the Brazilian Cerrado.

Analysis of the MCD64A1 product compared to the 5th validation approach shows a decrease in overall IoU and BA class performance from 2020 to 2022. The overall IoU for the 5th approach decreased from 0.867 in 2020 to 0.811 in 2021, and 0.798 in 2022, while the BA class IoU decreased from 0.464 in 2020 to 0.463 in 2021, and 0.301 in 2022. The most pronounced divergence occurred in 2022, with the BA class IoU 0.16 lower than in 2020 and 2021. Despite a similar pattern for the BA class, the 2020 and 2021 datasets differ in capturing NB areas, with IoU of 0.932 in 2020 and 0.895 in 2021, reflected in the higher overall IoU for 2020. CE analysis between MCD64A1 and the 5th approach shows lower CE in 2020 (0.227) and 2021 (0.282) compared to OE (0.461 and 0.432, respectively), indicating more BA misclassified as NB, resulting in more false negatives than false positives. In 2022, however, commission errors exceeded omission errors. Figure 14 presents the annual results.

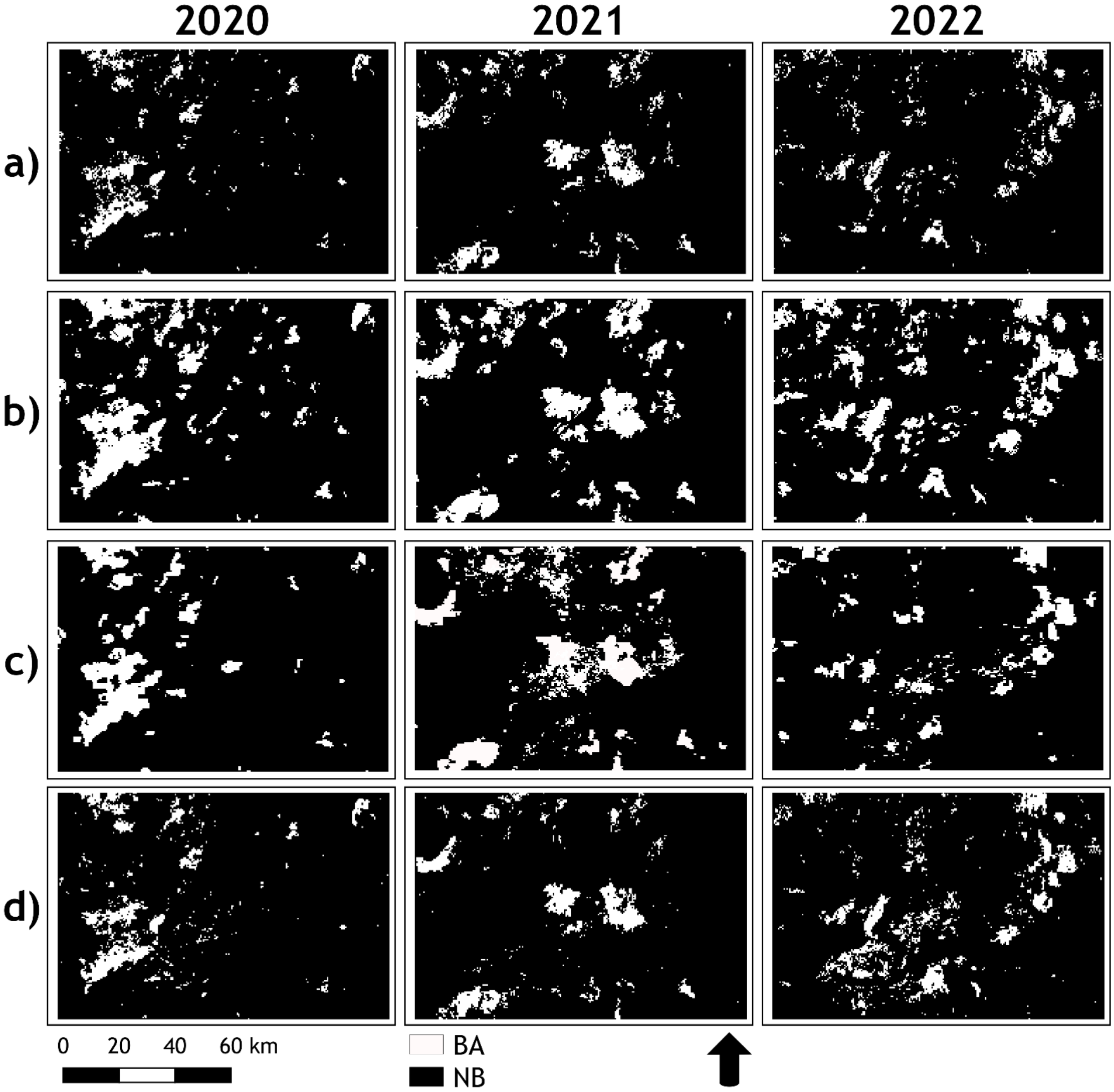

**Figure 14.** Maps used for comparisons: a) reference mapping containing TB as BA; b) reference mapping containing TB and PB as BA; c) MCD64A1 products; d) Results obtained applied the 5th validation approach.

Several factors may explain the lower performance of the 2022 dataset in detecting burned areas. The late inclusion of CBERS-4 imagery in 2022 likely played a key role, as historical data (prior to 2022) were unavailable in the INPE catalog. This limitation may have increased variability in the 2022 dataset, leading to misclassification of NB cells as BA. Radiometric differences between CBERS-4A, AMAZONIA-1, and CBERS-4 satellites likely contributed to this variability, though no major differences were observed in 2020 and 2021 results. The lack of a consistent, inter-annual time series with all three satellites since 2020 poses additional challenges to our analysis. Finally, differences in fire patterns, with smaller and more diffuse fires in 2022, may have influenced the classification process and the results.

Given the limitations discussed above, it is important to note that the dataset has not been fully explored, as only the BGR NIR bands were used, while other spectral indices are also available. Additionally, this study relies on the Random Forest algorithm, but the dataset could be further analyzed using other machine learning methods. Exploring alternative ML algorithms may help address some of the performance issues observed in 2022 and potentially improve the results for the other two years as well, providing more information about the effectiveness of WFI imagery for mapping burned areas.

## 4. Conclusions

Analyzing the usage of totally burned (TB), partially burned (PB), and non-burned (NB) cells, we observed lower efficiency in predicting partially burned cells due to increased spectral mixing within these areas. When comparing the classifier's performance across three input sets (1st, 2nd, and 3rd classification approaches), we achieved higher accuracy when TB cells were used to represent the burned area class, while PB and NB cells were combined to represent the non-burned class. Among the seven validation approaches based on the IoU metric, the 5th approach yielded the best overall results. This approach used TB to define the burned area class (BA), while PB and NB were merged to represent the non-burned class (NB) after classification had been performed with each label as an independent class. Our best results, in summary, were achieved by classifying the datasets with each label as an independent class. We then regrouped PB and NB cells into the NB class, while TB cells were kept as the burned area class.

The model's weaker performance in 2022, compared to other years, can be attributed to several factors warranting further investigation. While 2022 encompassed the most extensive time series, comprising 113 images, including those captured by the CBERS-4 satellite, the inclusion of CBERS-4 images in the INPE catalog only from May onwards resulted in a less comprehensive dataset for the earlier months. Furthermore, the utilization of three distinct satellites (CBERS-4, CBERS-4A, and AMAZONIA-1) introduced new radiometric data, potentially contributing to increased spectral variability. As noted by Oldoni (2022), this issue is challenging to assess comprehensively due to the absence of CBERS-4 imagery prior to May. Additionally, fire patterns and sample quality from 2022 may have influenced the performance of the RF algorithm, potentially introducing biases or overfitting that contributed to the observed variability. This is particularly significant, given that 2022 exhibited smaller overall burned areas but a higher prevalence of diffusely distributed burn patterns.

These observations are preliminary hypotheses that need further analysis. Despite the challenges, the dataset is a valuable resource for future research. Boosting algorithms, like AdaBoost, could improve model performance by combining weak learners into a strong model that corrects errors from the previous one. Support Vector Machines (SVMs) can work with both linearly and nonlinearly separable data, and could help clarify the dataset's structure. They find the optimal hyperplane to separate classes and use the kernel trick for non-linear cases to map the data into a higher-dimensional space for linear separation. The dataset can also be processed using deep learning architectures like Long-Short Term Memory (LSTM) networks to account for temporal dynamics. LSTM networks capture short and long term patterns, allowing analysis of fire behavior over different time intervals.

Future tests could help researchers develop automated methods to integrate WFI images into the DETER project’s daily operations, supporting photo interpreters, optimizing near real-time analyses, and providing reliable data for law enforcement. If WFI images from the three satellites can be combined to systematically map burned areas in the Cerrado, they could be incorporated into BDQueimadas—INPE's program, a database containing hotspots and burned area data for end users. The present methodology could also be investigated to other fire-prone regions using sensors operating in the visible and NIR bands with similar temporal resolution. However, if the WFI sensor includes a shortwave infrared (SWIR) band, the methodology might need to be revised, as SWIR enhances the spectral detection of burned areas, potentially improving accuracy and expanding its applications.

In this article, we examine the performance of the dataset with different classification and validation approaches, focusing on data derived from the BGR NIR bands. We begin by analyzing different combinations of the TB, PB, and NB labels to define the “burned area” and “unburned area” classes on an annual basis. Additionally, other temporal approaches and the use of spectral indices can be explored for further insights.

**Disclosure statement**
No potential conflict of interest was reported by the author(s).

**Data availability statement**
The data that support the findings of this study are openly available in Science Data Bank(ScienceDB) at https://doi.org/10.57760/sciencedb.14040.

**Funding details**
The authors thank the São Paulo Research Foundation (FAPESP, grant no 2023/09118-6), and the Brazilian National Council for Scientific and Technological Development (CNPq, grant no 302205/2023-3).